\documentclass[11pt,a4paper]{article}
\usepackage[hyperref]{acl2021}
\usepackage{times}
\usepackage{latexsym}
\usepackage{float}
\usepackage{graphicx}% http://ctan.org/pkg/graphicx

 \usepackage{graphicx}
\usepackage{microtype}

\aclfinalcopy % Uncomment this line for the final submission
\title{A Sequence to Sequence Model for Extracting Multiple Product Name Entities from Dialog}

\author{Praneeth Gubbala \\
  Walmart Global Tech \\
  \texttt{p0g00i0@wal-mart.com} \\\And
  Xuan Zhang \\
  Walmart Global Tech \\
  \texttt{xuancs@vt.edu} \\}

\date{}

\begin{document}
\maketitle
\begin{abstract}
E-commerce voice ordering systems need to recognize multiple product name entities from ordering utterances. 
Existing voice ordering systems such as Amazon Alexa can capture only a single product name entity. 
This restrains users from ordering multiple items with one utterance.
In recent years, pre-trained language models, e.g., BERT and GPT-2, have shown promising results on NLP benchmarks like Super-GLUE. 
However, they can’t perfectly generalize to this Multiple Product Name Entity Recognition (MPNER) task due to the ambiguity in voice ordering utterances. 
To fill this research gap, we propose Entity Transformer (ET) neural network architectures which recognize up to 10 items in an utterance. 
In our evaluation, the best ET model (conveRT + ngram + ET) has a performance improvement of 12\% on our test set compared to the non-neural model, and outperforms BERT with ET as well. 
This helps customers finalize their shopping cart via voice dialog, which improves shopping efficiency and experience.
\end{abstract}

\section{Introduction}
% \begin{figure*}[ht]
%       \includegraphics[width=1.0\textwidth]{voice-ordering-ecnlp2021/pics/SampleUtterance.png}
%     \centering
%     \caption{A Sample Utterance}
%     \label{fig:SampleUtterance}
% \end{figure*}

Take a sample utterance like  ``add seven apples one gallon of milk two bags of sunflower seeds fresh garlic disposable wipes” as an example. %(Figure ~\ref{fig:SampleUtterance}). 
The task of product name entity recognition is to identify product entities such as ``apples", ``milk", ``sunflower seeds", ``fresh garlic", and ``disposable wipes".
Both the amount and diversity of product name are challenges for segmenting and extracting them.
Usually, the Automated Speech Recognition (ASR) systems don't produce punctuations and symbols when translating voice into text, due to their limitation.
This significantly increases the challenge to separate adjacent product names.
Since these barriers (e.g., entity amount, missing punctuation, and product diversity) are unusual for traditional NER tasks, existing methods may not generalize to MPNER well.

\section{Approach}

\subsection{Model Architecture} 

\begin{figure*}
      \includegraphics[width=1.0\textwidth]{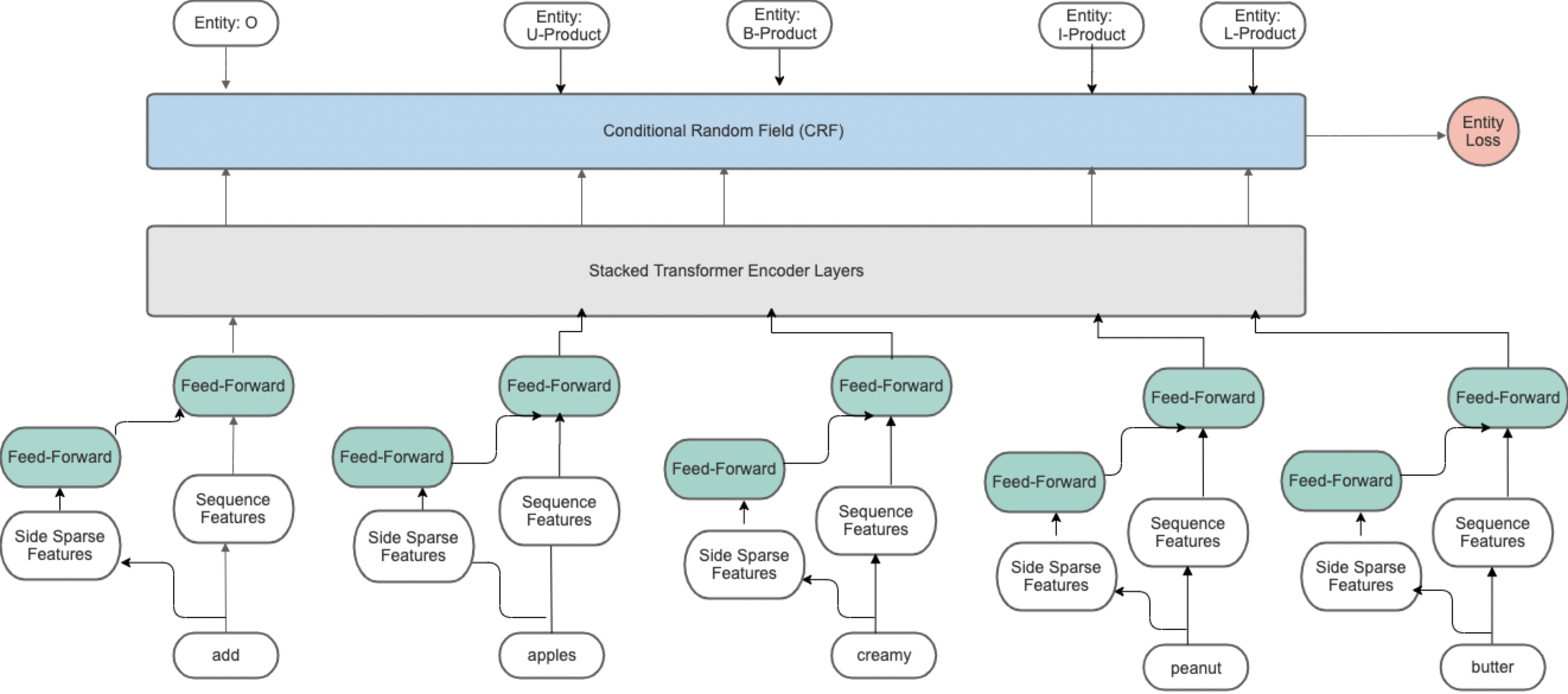}
    \centering
    \caption{Entity Transformer architecture}
    \label{fig:architecture}
\end{figure*}

An Encoder-Decoder architecture, called Entity Transformer (ET), is developed for the proposed model inspired by \cite{diet}, shown in Figure~\ref{fig:architecture}. 
In this design, the input sequence are encoded by stack of transformer layers \cite{vaswani2017attention}. The encoder maps input sequence ($x_1,....,x_n$) to a sequence of  continuous representations E= ($e_1,...,e_n$).
From E, the Conditional Random Field (CRF) decoder to produce the sequence of output entity tag labels predictions ($l_1,...,l_n$).

\begin{table*}
\centering
\caption{ Model performance on MPNER task.}
\begin{center}
\begin{tabular}{l | l | l | l}
 Sparse Features &  Dense Features & Training F1 &  Test F1 \\
\hline
count vectors (word + char n-grams(n $\leq$ 4)) & conveRT  &  \textbf{99.42}   & \textbf{85.84} \\
count vectors (word + char n-grams(n $\leq$ 4)) + lexical & conveRT     &  99.23   & 85.83 \\
count vectors (word + char n-grams(n $\leq$ 4))  + lexical  & BERT     &  99.18   & 84.89 \\
count vectors (word + char n-grams(n $\leq$ 4))  & Roberta     &  99.33   & 77.09 \\
count vectors (word + char n-grams(n $\leq$ 4)) + lexical &   Not Present     &  99.39   & 80.58 \\
ensemble (crf , svm ) & Custom fastext & 94.30 & 73.10 \\
\end{tabular}
\end{center}
\label{tab:ModelPerformance}
\end{table*}

\subsubsection{Features}
For dense features, we extracted input sequence features from the various pre-trained language models \cite{hugging-face},  such as conveRT \cite{henderson2020convert} from PolyAI, BERT\cite{devlin-etal-2019-bert}, and Roberta \cite{RoBERTa}.
Side Sparse features are one-hot token level encodings and multi-hot character n-gram ({n$\leq$4}) features. The sparse features are passed to the feed-forward layer whose weights are shared through the input sequence. The Feed-forward Neural Network (FNN) layer output and dense sequence features are concatenated before passing to Transformer encoder layers. 
\subsubsection{Transformers}
For encoding the input sequences, we used stacked transformers layers (N $\leq$\textbf{6}) with relative position attention. Each transformer encoder layer composed of multi-headed attention layers and point wise feed forward layers. These sub layers produces a output of dimension $d_{model} = 256$.The number of attention heads: $N_{heads}=4$.The number of units in transformer: $S=256$.The number of transformer layers: $L= 2$.
\subsubsection{Conditional Random Field}
The decoder for named entity recognition task is Conditional Random Field (CRF) \cite{lample2016neural} which jointly models the sequence of tagging decisions of an input sequence. we used BILOU tagging scheme \cite{ratinov-roth-2009-design}.
\subsection{Model Training and Inference}
% \subsubsection{Optimizer}
We used Adam optimizer \cite{kingma2017adam} with the initial learning rate of 0.001. 
% \subsubsection{Batching}
As NER labels are at the token level, we didn't apply any batching strategy on the training data. 
We increase batch size throughout training progress as a source of regularization from 64 to 256 \cite{smith2018dont}. 
The ET model was developed with Tensorflow.
The model training time on various hardware can be found in Table~\ref{tab:training-time}.The inference time of ET models is 80 ms in average for one voice reorder utterance.

\begin{table}
\centering
\caption{ Model training time comparison: train (conveRT + ngram + ET) model for 100 epochs}
\begin{center}
\begin{tabular}{c | c | c | c}
 CPU Core & GPU &  RAM & Training Time   \\
 \hline
 8 & - &  64 GB  & 36 hours  \\
 16 & - &  56 GB  & 18 hours  \\
6 & 1 x K80 &  56 GB  & 12 hours \\
6 & 1 x V100 & 112 GB & 7 hours \\
\end{tabular}
\end{center}
\label{tab:training-time}
\end{table}

\section{Evaluation}
We created a MPNER data set consists of about 1M product name entities from voice order shopping utterances. This dataset has a 500k training subset and a 65k test subset utterances.The utterance in each data example has a random number of (between 1 and 10) products. This data is generated from 70 seed voice order utterance variations, using synonym-based data augmentation of the most popular 40k products from various departments. The test data created with 5k unseen products also from the same categories.
Table~\ref{tab:ModelPerformance} shows the model evaluation results on the MPNER data set. Our best model configuration is with sparse features of word count vectors, char n-gram count vectors (n$\leq$4), and conveRT \cite{henderson2020convert} pre-trained dense embeddings.

% \begin{table*}
% \centering
% \caption{ Model performance on MPNER task.}
% \begin{center}
% \begin{tabular}{l | l | l | l}
%  Sparse Features &  Dense Features & Training F1 &  Test F1 \\
% \hline
% \textbf{count vectors} & \textbf{conveRT} & \textbf{99.42} & \textbf{85.84} \\
% count vectors  + lexical & conveRT     &  99.23   & 85.83 \\
% count vectors  + lexical  & BERT     &  99.18   & 84.89 \\
% count vectors   & Roberta     &  99.33   & 77.09 \\
% count vectors + lexical &   Not Present     &  99.39   & 80.58 \\
% ensemble (crf , svm ) & Custom fastext & 94.30 & 73.10 \\
% \end{tabular}
% \end{center}
% \label{tab:ModelPerformance}
% \end{table*}

In the future, we plan to investigate customized pre-trained models for MPNER task.
% \input{SEC4-Conclusion}

% \section*{Acknowledgments}

% The acknowledgments should go immediately before the references. Do not number the acknowledgments section.
% \textbf{Do not include this section when submitting your paper for review.}

\newpage

\bibliographystyle{acl_natbib}
\bibliography{voice_order}

%\appendix

\end{document}